\documentclass[a4paper, 11pt, conference]{ieeeconf}      
\IEEEoverridecommandlockouts                              

\usepackage{lipsum}
\usepackage{graphicx}                       
\usepackage{stfloats}
\usepackage[tight,footnotesize]{subfigure}  

\usepackage{amssymb,amsmath}
\usepackage{textcomp}
\usepackage{mdwmath}
\usepackage{mdwtab}
\usepackage{eqparbox}
\usepackage{rotating}                       
\usepackage{array}                          
\usepackage{listings}                       
\usepackage[ruled,vlined,noresetcount]{algorithm2e}
\usepackage{listings}
\usepackage[noend]{algpseudocode}
\usepackage{bm}
\usepackage[colorlinks,
            linkcolor=blue,
            anchorcolor=blue,
            citecolor=blue,
            bookmarks=false
            ]{hyperref}
\usepackage{cite}

\usepackage{amsmath}
\usepackage{amssymb}
\usepackage{eucal}
\usepackage{graphicx}
\usepackage{float}
\usepackage{subfigure}
\usepackage{amsmath}
\usepackage{algpseudocode}
\usepackage{stfloats}
\usepackage{multirow} 
\usepackage{xcolor}
\usepackage[ruled,vlined]{algorithm2e}
\usepackage{amssymb}
\usepackage{booktabs}

\title{\LARGE \bf A LiDAR-Inertial-Visual SLAM System with Loop Detection}

\author{Kangcheng Liu}  

\begin{document}

\maketitle
\thispagestyle{empty}
\pagestyle{empty}

\begin{abstract}
We have proposed, to the best of our knowledge, the first-of-its-kind LiDAR-Inertial-Visual-Fused simultaneous localization and mapping (SLAM) system with a strong place recognition capacity. Our proposed SLAM system is consist of visual-inertial odometry (VIO) and LiDAR inertial odometry (LIO) subsystems. We propose the LIO subsystem utilizing the measurement from the LiDAR and the inertial sensors to build the local odometry map, and propose the VIO subsystem which takes in the visual information to construct the 2D-3D associated map. Then, we propose an iterative Kalman Filter-based optimization function to optimize the local project-based 2D-to-3D photo-metric error between the projected image pixels and the local 3D points to make the robust 2D-3D alignment. Finally, we have also proposed the back-end pose graph global optimization and the elaborately designed loop closure detection network to improve the accuracy of the whole SLAM system. Extensive experiments deployed on the UGV in complicated real-world circumstances demonstrate that our proposed LiDAR-Visual-Inertial localization system outperforms the current state-of-the-art in terms of accuracy, efficiency, and robustness.

\end{abstract}

\section{Introduction}
Currently, 3D localization becomes an essential part and prerequisite in autonomous robotic systems. In the past years, Light Detection And Ranging Sensors (LiDAR) have been widely deployed in various robotics platforms such as automatically navigated aerial robots and ground robots \cite{liu2022industrial}. The reason is that LiDAR has high accuracy in 3D measurement and great robustness to the fast movement of robots. The most popular strategy for large-scale robotics mapping is LiDAR-Inertial Odometry, and it has witnessed great success in low-speed and simple robotic navigation scenarios when the lighting conditions are well-controlled. However, when faced with the reflective materials such as glasses, and the long corridor circumstances without any prominent geometric structures, the LIO system suffers from noises and low accuracy. For currently widely adopted solid-state LiDARs with a limited field of view, the problems become more prominent. However, the LiDAR is still the most popular and prominent sensor because it is the one of the most fast and accurate robotic sensors \cite{liu2017avoiding, liu2020fg, liu2022weak, liu2021fg, liu2019deep, liu2022fg, zhao2021legacy, liu2022ARM, liu2022CYBER2, liu2022ICCA1, liu2022ICCA2,liu2022CYBER1}. It has great potential to be applied to industrial building inspections. 

\begin{figure}[tbp!]
\centering
\includegraphics[scale=0.21]{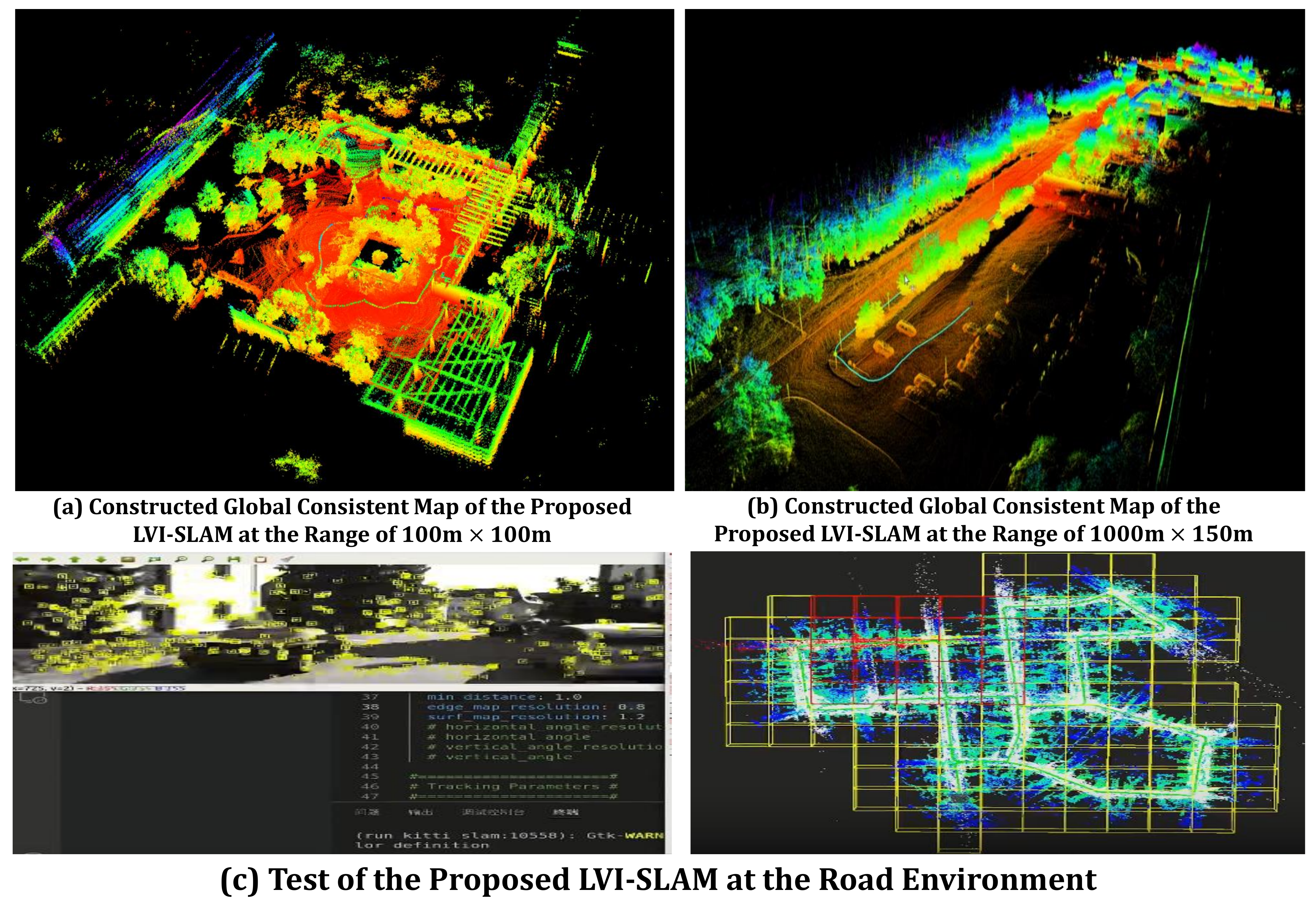}
\caption{Testing of our proposed LVI-SLAM system at various of real-site test circumstances. We have proposed a general LiDAR-Visual-Inertial Localization and Mapping System that is both appropriate for Mechanical LiDAR  and Solid-State LiDAR. The effectiveness of the proposed LVI-SLAM system is tested extensively for both indoor and outdoor environments.}
\vspace{-3.9mm}
\label{fig_overall}
\end{figure}
Motivated by the challenges of large-scale LiDAR mapping, in recent years, the loosely coupled LiDAR-Inertial-Visual Fusion SLAM (LVI-SLAM) frameworks have been developed to improve the robustness and effectiveness of the LVI-SLAM system. The Visual-LOAM is the pioneering work that employs a loosely coupled VIO system for the initialization of the LiDAR odometry and mapping system. A solution has also been proposed for 3D heritage modeling with satisfactory performances in 3D reconstructions. However, it merely performs well in building reconstruction circumstances and has limited performance in the real-time mapping of the environments. The unified multi-modal landmark tracking methods for the mobile platform have been proposed to improve the accuracy of the LiDAR-Visual-Inertial odometry performance in circumstances with long corridors and dark environments. However, the methods are just simple combinations of the existing State-of-the-arts (SOTAs) localization and mapping systems. Also, the effectiveness under high-speed movements remains to be demonstrated.  

Recently, tightly coupled LVI-SLAM frameworks have been proposed. It takes the IMU measurements, the sparse visual features, and the LiDAR point clouds feature as input. The multi-state constrained Kalman Filter (MSCKF) is utilized to make a real-time fusion of the above measurements with online spatial and temporal calibrations. To improve the robustness and the effectiveness of the first version, the LIC-Fusion SLAM framework has been proposed, It adopts a point clouds planar feature tracking method across diverse LiDAR scans using the sliding window-based approaches. The poses are also refined within the sliding window. The LVI-SAM~\cite{liu2022ICCA2, liu2022ws3d, yuzhi2020legacy, liu2022robustcyber, liu2022semicyber, liu2022lightarm} framework is based on the traditional LOAM pipeline, which supports both the joint optimizations of the LiDAR odometry and the visual odometry and the individual optimizations of the VIO and the LIO system. The individual working mechanisms of LIO and VIO are activated when a failure occurs in one of the systems. In summary, the tightly-coupled LVI-SLAM systems have better accuracy and robustness but heavier computational costs. A more robust and efficient LVI-SLAM system requires to be developed. 

In this work, we have proposed a robust LiDAR-Inertial-Visual Fusion framework with great robustness and efficiency. The integrated system has robust performance when faced with a noisy point cloud environment and fast-moving camera sensors.  In summary, we have proposed the first-of-its-kind LiDAR-Inertial-Visual Fusion SLAM system~\textbf{LVI-SLAM}, which has the following three prominent contributions:
\begin{enumerate}
    \item  Firstly, we have proposed a tightly coupled LiDAR-Visual-Inertial fusion SLAM (LVI-SLAM) framework consisting of an LIO system for the 3D point clouds-based mapping and a VIO system for establishing the 2D-3D visual alignment. 
    
     \item  Secondly, we have proposed to project the images obtained in the VIO system into 3D to compensate for the optimization errors caused by long-term drifting of LiDAR odometry, which boosts the whole localization accuracy and also the final mapping performances. Our proposed VIO system does not rely on the extraction of the keypoints in the front-end of the odometry and it merely relies on the 2D-3D projection, which improves the robustness of the visual SLAM system in environments lacking textures. 
     
     \item Thirdly, we have proposed the front-end iterative Kalman Filter-based odometry for the tightly coupled front-end optimization of our LVI-SLAM system. We have also proposed factor graph-based optimizations for the backend optimizations of our LVI-SLAM system.
     
    \item  Finally, we have proposed the loop closure detection network, termed \textit{LC-Net}, to further enhance the performance of the proposed SLAM system. Compared with many previous methods which are merely demonstrated by the simulations and simple application cases, we have done extensive real-site experiments in the indoor, outdoor, and underground tunnel subterranean environments to demonstrate the effectiveness of our proposed method.
\end{enumerate}
\begin{figure*}[ht]
\centering
\includegraphics[scale=0.369]{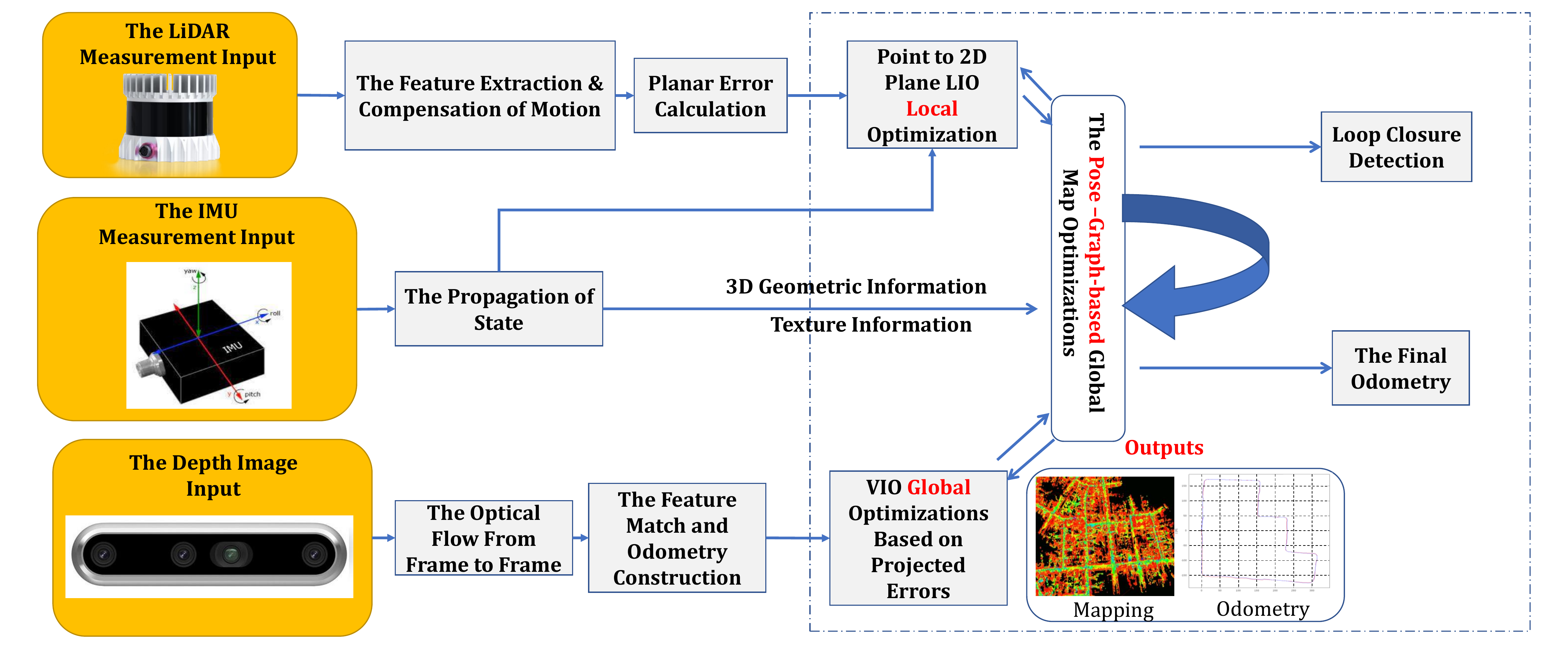}
\caption{The final framework of our proposed tightly coupled LVI-SLAM system. It fuses the measurement from the three kinds of sensors including the LiDAR sensor, the camera sensor, and the inertial sensor. We have proposed optimization functions for both the front-end odometry and the back-end optimizations.}
\label{fig_livo}
\vspace{-5mm}
\end{figure*}

 \section{The LVI-SLAM System structure} 
 The overall system framework of our proposed LiDAR-Visual-Inertial Fused SLAM system is shown in Fig. \ref{fig_livo}. The point clouds from the LiDAR sensors and the images from the camera sensors are taken as input to our VIO system and the LIO system, respectively. In the LIO system, we firstly eliminate the motion variance by backwards propagation from the IMU sensors. Then the point-to-2D-plane residual is calculated for the optimizations of the global point cloud map in the front-end. In a similar way, the proposed VIO system uses the frame-to-map 2D photo-metric errors to formulate a 2D-image based optimization in the front-end. The point-to-plane error in the LIO system and the 2D frame to map errors are all tightly coupled into the final factor graph optimizations to obtain the global consistent 3D map. 



\subsection{The LiDAR Scan-to-Map Measurement}
The LiDAR-based odometry can be viewed as the process of solving the ego-motion of the robot platform through comparison and association of the input consecutive LiDAR scans. While we are registering the scanned points to the map, we assume that each point $\textbf{p}_i$ is within the neighbouring plane with a normal $\textbf{n}_j$ and a center point $\textbf{q}_c$. If transforming the measured point clouds in the LiDAR-based local frame to the global frame utilizing the ground truth state $x_k$, the residual error ought to be zero, which can be represented as:
\begin{equation}
    \textbf{r}_{l}(\textbf{x}_k, \textbf{p}_i)=\textbf{n}_j (\textbf{T}_L \textbf{T}_C \textbf{p}_i- \textbf{q}_c)=\textbf{0}
    \label{lio_optim}
\end{equation}
Denote the extrinsic of the LiDAR frame with respect to the IMU frame as $\textbf{T}_L$, and the extrinstics of the camera frame with respect to the IMU frame as $\textbf{T}_C$. The Eq. \ref{lio_optim} above can serve as the front end optimization function for our LIO-subsystem.

\subsection{The Sparse Direct Visual Alignment Measurement Model}
\begin{figure}[t]
\centering
\includegraphics[scale=0.36]{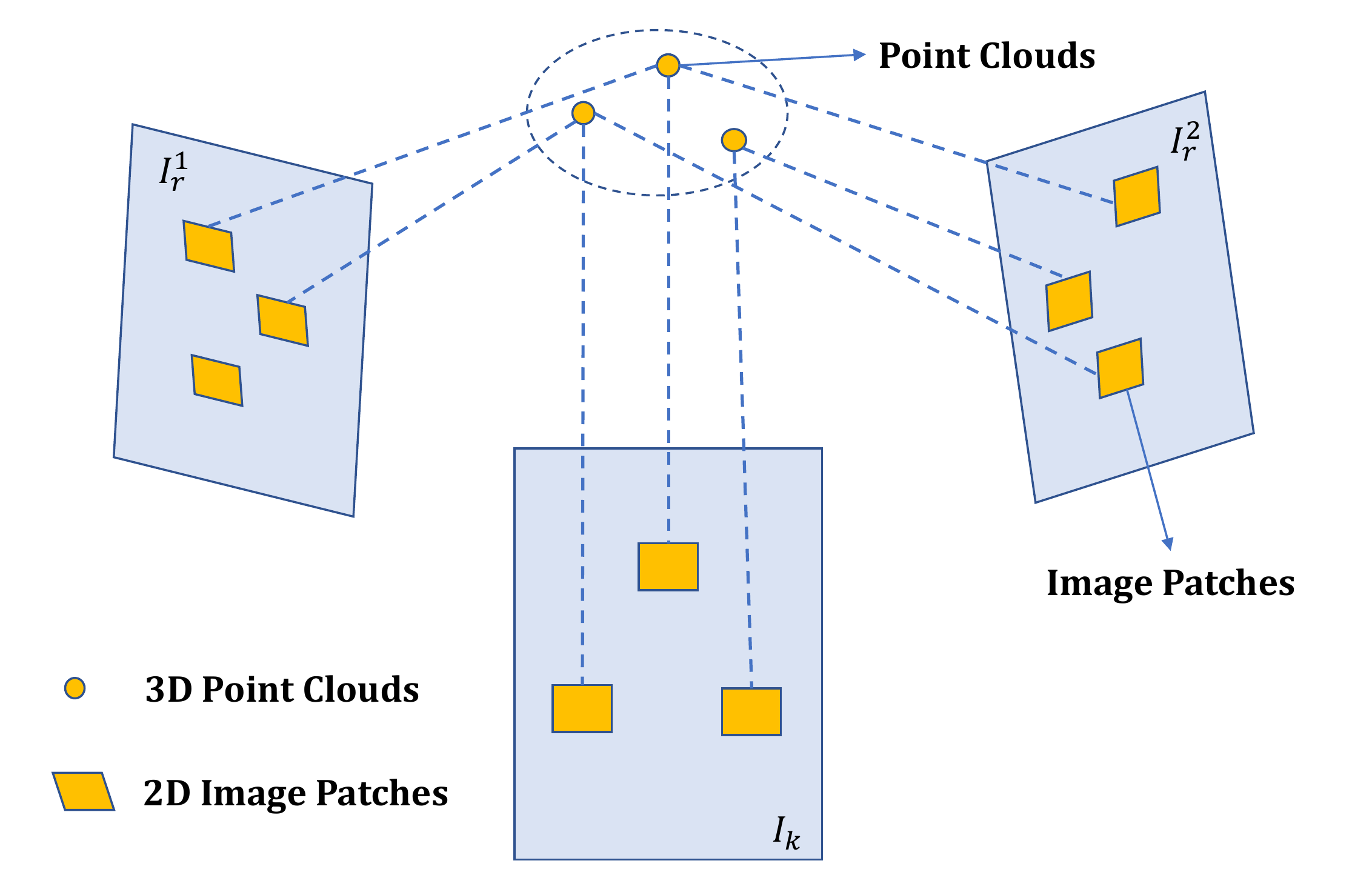}
\caption{Transforming the pose of the current frame $I_k$ in a global manner to minimize the photo-metric error between the projected 3D point clouds from the image patches to minimize the photo-metric error.}
\label{fig_camera_frame}
\vspace{-3mm}
\end{figure}

As shown in Fig. \ref{fig_camera_frame}, if an image is received at the time $t_k$, we can extract the mapped corresponding 3D points $\textbf{p}_i$ that fall within the current field of view. As shown in Fig. \ref{fig_camera_frame}, for each point, we will select the path in images that can see the points from the nearest observation angle with the current 2D image as the reference, which is represented as $\textbf{Q}_i$. Finally, converting the mapping point $p_m$ to the current 2D image $S_k$ with the true state, the final photo-metric error should be made to zero, which can be represented as follows:
\begin{equation}
    \textbf{0}= \textbf{r}_c(\textbf{x}_k, \textbf{p}_i)= S_k(\mathcal{L}(\pi (\textbf{T}_C\textbf{T}_L\textbf{p}_i)))- \textbf{Q}_i
\end{equation}
where the $\mathcal{L}$ is the camera projection pinhole model. As shown in Fig. \ref{fig_camera_frame}, through this kind of alignment, the 2D-3D correspondence can be established. And the final photo-metric error is formulated into the iterative Kalman Filter  
\subsection{The Front-End Iterative Kalman Filter-based Odometry}

Combining the LiDAR odometry-based measurement and the visual odometry-based measurement, we can obtain the final maximum posterior probability based state estimation, which can be formulated as the summed optimization function as follows:

\begin{equation}
\mathop{argmin} \limits_{\sigma} \sum_{j=1}^{l} \| \textbf{r}_{l}(\textbf{x}_k, \textbf{p}_i)\|+ \sum_{j=1}^{c} \| \textbf{r}_{c}(\textbf{x}_k, \textbf{p}_i)\|
\end{equation}
The optimization in the equation above is non-convex and can be iteratively solved by the Gauss-Newton method. The optimization can be simplified into an iterative Kalman filter. In this way, the visual 2D pixel information and the LiDAR-based 3D point cloud information are both taken into consideration for the front-end 3D point cloud map optimization. 

\subsection{The Back-End Factor-Graph-based Global Optimization for SLAM}
In this Subsection, we propose the complete framework for our backend factor-graph-based SLAM system. In most applications, we need to derive the motion of the robot in 3D space. Take the motion in the 2D space as an example, in those situations, we need to consider the complete six-degree-of-freedom information which combines the position and the orientation. Specifically, by combining position and orientation $SO(2)$, we can obtain a special Euclidean group $SE(2)$, then the pose of the robot $x_i$ can be represented in the $SE(2)$.
In most industrial situations, the object is the unmanned aerial vehicle or the ground vehicle which requires movement in the uneven ground. Therefore, the pose in the most robotic applications is $x_i \in SE(3)$, which is a six-dimensional manifold. In the situations of $SE(2)$ and $SE(3)$, we often consider the rate of motion. The rate of motion is multiplied with a finite amount of time $\Delta \tau$ to obtain a motion increment $\sigma$. Specifically, we define the angular velocity, the linear velocity, and the increment $\sigma$ respectively:

\begin{equation}
\sigma = \begin{bmatrix} \omega \\ v \end{bmatrix} \Delta  \tau
\end{equation}
Assume $z \in \mathbb{R}^m$ is the known measurement vector, we can consider a simple minimum problem on a single pose $x$, which is formulated as:

\begin{equation}
    \textbf{x}^{\star}= argmin \|h(\textbf{x})-z\|^2_{\Sigma}
\end{equation}

To minimize the objective function, we need to represent how the nonlinear measurement function $h(\textbf{x})$ operates at the neighbourhood of the basic pose $\textbf{x}_{0}$. Therefore, we need to calculated the Jacobian matrix $\textbf{H}_0 \in \mathbb{R}^{m\times 6}$. 
\begin{equation}
h(\textbf{x}_0+\bm{\sigma}) \approx h(\textbf{x}_0) + \textbf{H}_0\bm{\sigma}
\label{eq_jaco}
\end{equation}
Note that $\bm{\sigma} \in \mathbb{R}^6$. The vector $\bm{\sigma}$ is six-dimensional, which represents all the movable directions in the six-dimentional maniford. The numerical differentiation, the symbolic differentiation or the automatic differentiation can be used in solving Eq. \ref{eq_jaco}. It can give the final solution to  $\textbf{H}_0$.

As long as we have the approximated formula (Eq. \ref{eq_jaco}), we can minimize the objective function with respect to the local coordinates. 

\begin{equation}
\bm{\sigma^{\star}}= \mathop{argmin}\limits_{\sigma}\| h(\bm{x}_0+\bm{H_0}\bm{\sigma}-\bm{z})\|^{2}_{\sum}
\label{eq_jacglo}
\end{equation}
In the final pose optimizations-based SLAM systems, we do not merely want to restore a single pose. Instead, we want to restore all the poses throughout the whole trajectory. Generally speaking, the pose graph-based SLAM involves two types of factors. The first is the unitary factor which is the absolute measurement from the pose prior (such as those obtained from GPS information). The second is the binary factor, which is the relative pose constraints obtained from the LiDAR sensors.

\begin{figure}[t]
\centering
\includegraphics[scale=0.5]{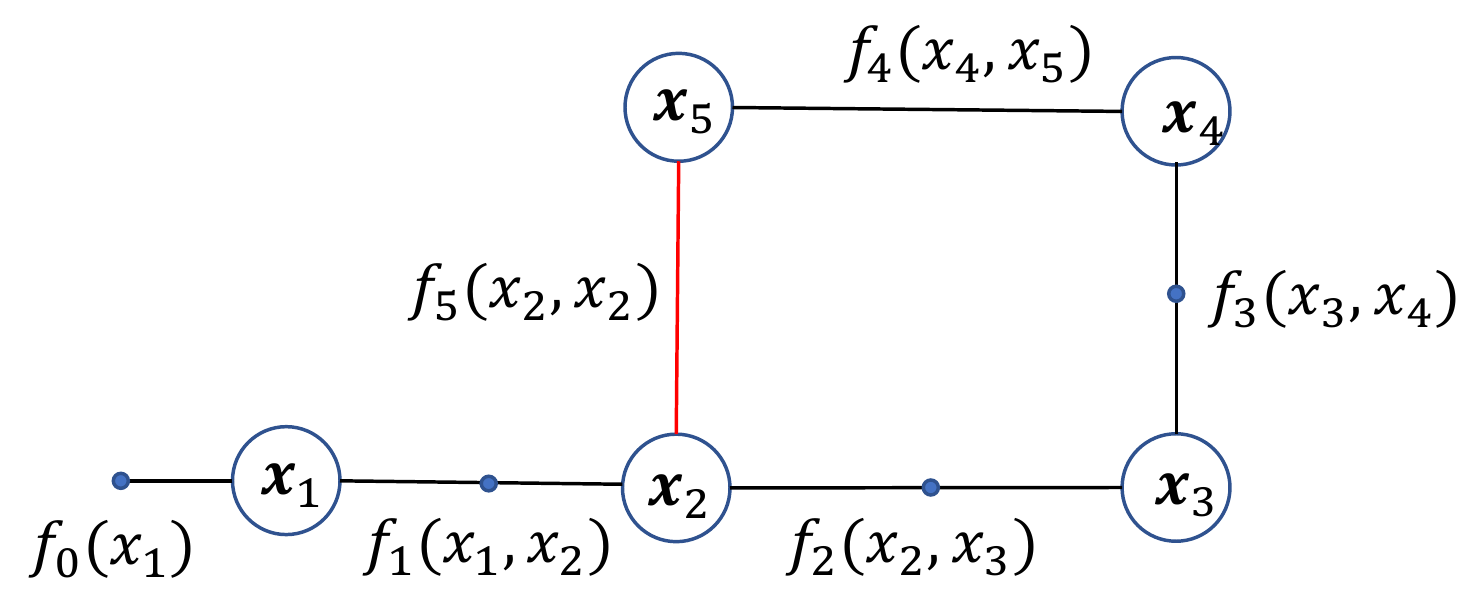}
\caption{The factor graph of the proposed LiDAR-Inertial-Visual fused SLAM system}
\label{fig_factorgraph}
\end{figure}

We have shown a simple factor graph of SLAM system in Fig. \ref{fig_factorgraph}. In order to anchor the factor graph, we have added the unary factor $f_{0}(\textbf{x}_1)$. In addition, with the traversing of the robot within the environment, the binary factor $f_t(\textbf{x}_t, \textbf{x}_{t+1})$ is corresponding to the generated odometry information. Finally, to perform the pose graph optimizations, we solve the local coordinates of all the poses by minimizing the following linearized observation factor. 

\begin{equation}
\begin{aligned}
\bm{F}(pose)= \mathop{argmin} \sum_{i}\|h(\bm{x}_i+\bm{H_i}\bm{\sigma}-\bm{z})\|^{2}_{\sum} \\
+ \|g(\bm{x_i},\bm{x_j})+\bm{F}_i\bm{\sigma}_j+\bm{G}_j\bm{\sigma}_j-\bm{z}\|^2_{\sum}
\label{eq_pose_glo}
\end{aligned}
\end{equation}

The $\bm{F}(pose)$ is the sum of all the incremental pose coordinates. $h$ and $g$ is the corresponding unary and binary measurement equations. The $\bm{H}_i$, $\bm{F}_i$ and $\bm{G}_j$ are the corresponding Jacobian matrices. According to our experiments, the proposed optimization function can provide better performance in localization accuracy compared with other global optimization methods. 
\section{The Learning based Loop Closure}
\begin{figure}[t]
\centering
\includegraphics[scale=0.060]{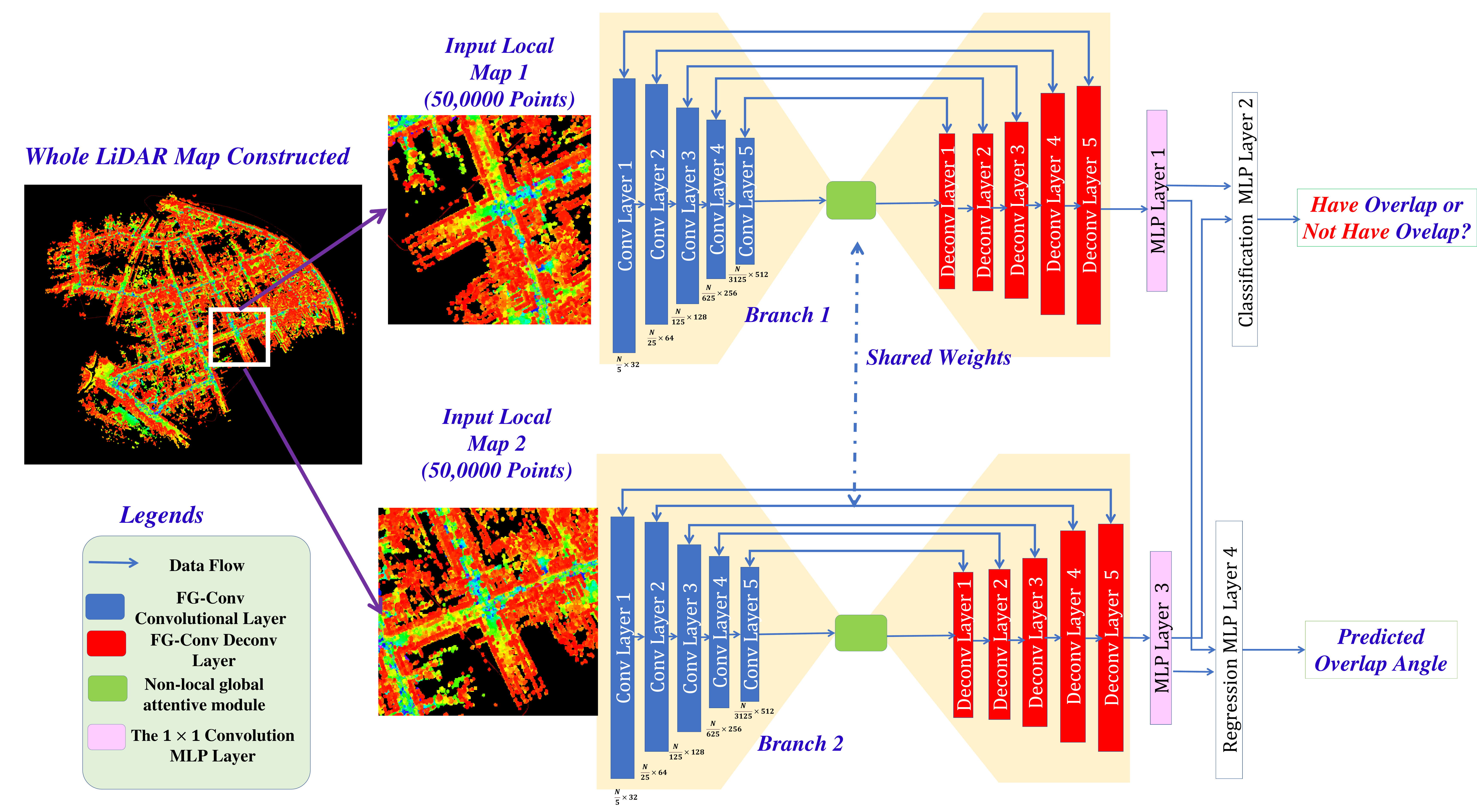}
\caption{Overall framework of our proposed Loop Closure Network (\textit{LC-Net}). We propose to input the LiDAR point clouds to the encoder decoder based network to give the prediction that if the two LiDAR scans have overlap between each other. Also, we regress the overlap angle between those two scans.}
\label{fig_loop_clo}
\end{figure}
Also, we have proposed the learning based loop closure strategy, which is termed as Loop Closure Network (\textit{LC-Net}), as shown in the Fig. \ref{fig_loop_clo}. The learning based loop closure aims at judging whether the current LiDAR scan is the same candidate scan of the same place that has been previously visited. The loop closure is very significant in large-scale mapping because it can correct the accumulative drifts when doing incremental mapping, especially in specific large-scale complicated environments of more than $10^4$ square meters. 
\subsection{General Architecture} 
The learning-based approaches have been widely adopted and applied in 3D vision \cite{liu2022datasets1, yang2022datasets, liu2022d, liu2022integrated, liu2022light, liu2022rm3d, liu2022enhanced, liu2022efficient}.
We utilize the effective FG-Conv \cite{liu2021fg} as our proposed network backbone for its high efficiency. It has been optimized elaborately for real-time applications and can realize a very fast inference speed. We have adopted two streams of FG-Conv as the siamese network. The network adopted an encoder-decoder-based structure with skip connections. In each stage of our network, we have down-sampled the number of point clouds by five times. In the earlier stages of the network architecture, we utilized random sampling because the number of point clouds in large. In the deep stages of the network architecture, we have adopted learning-based sampling to focus more on the important feature that matters most to loop closure detection. In the convolutional layer, we combine the advantages of point based branch and the convolution-based branch to improve the accuracy of our proposed approach. We have utilized merely 6\% of data for training, so our approach is also data efficient. It is demonstrated our network has great domain adaptation capacity and our network is trained on SemanticKITTI and tested on other various complicated 3D autonomous driving circumstances.   
\vspace{-3mm}
\subsection{Loss Function and Optimizations}

\begin{figure}[t]
\centering
\includegraphics[scale=0.25]{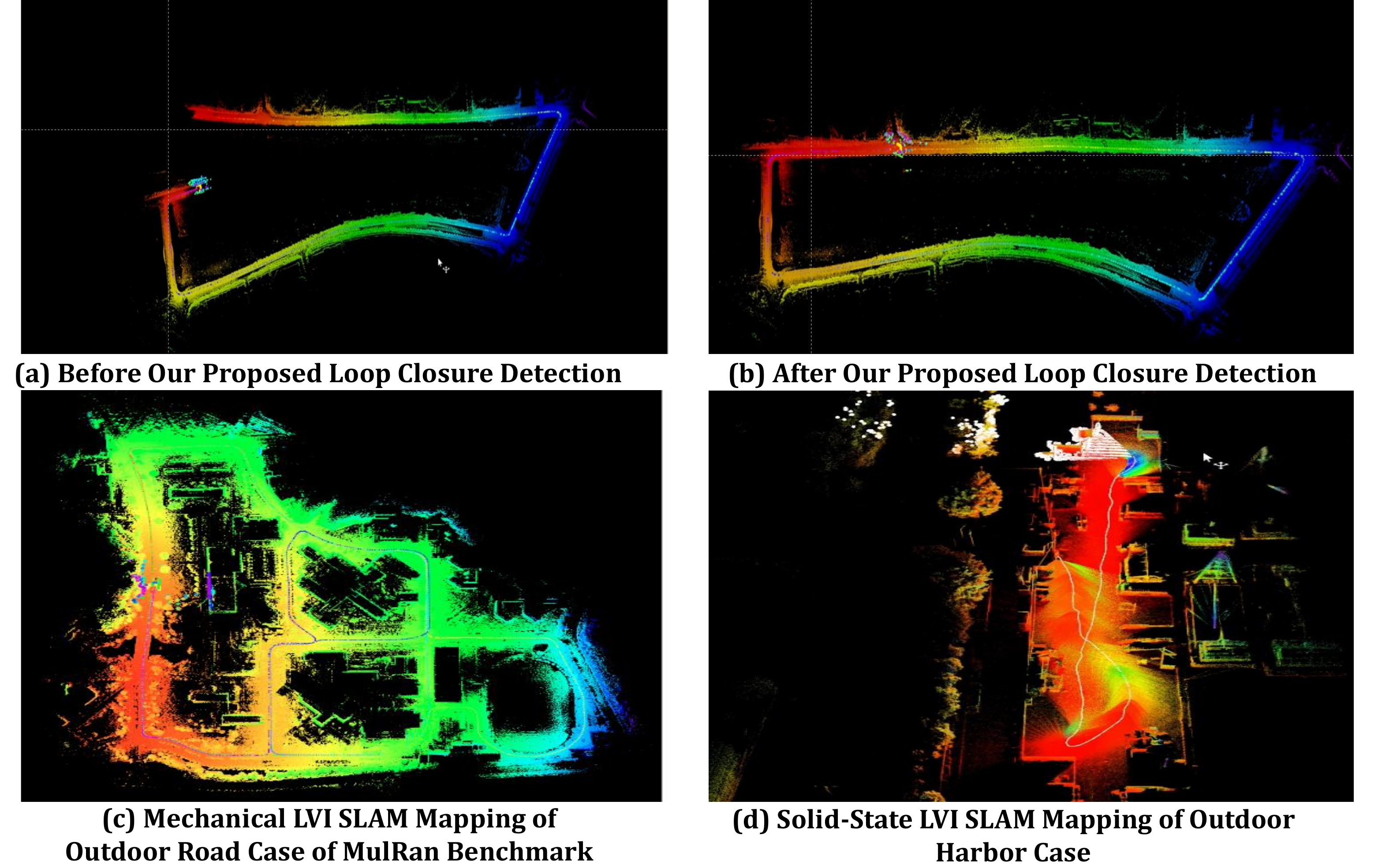}
\caption{The final demos of the Proposed LVI-SLAM mapping results. It can be seen that on Mulran benchmark, our proposed learning-based loop closure detection method can achieve satisfactory accuracy and correct large error (subfigure (a) which is the global map before proposed loop closure detection method) to a global consistent map (subfigure (b) which is the map obtained after we do the loop closure.) The Subfigure (c) and Subfigure (d) demonstrate our testing results on the outdoor road case and outdoor harbor case respectively. It can be seen that the global consistently mapping can be achieved both in the indoor and outdoor circumstances with mechanical and solid-state LiDAR. The robustness of our proposed approach is demonstrated. Also, we have tested our proposed LVI-SLAM system at the Hong Kong city environment as shown in the Subfigure (e) and Subfigure (f).}
\label{fig_exp}
\vspace{-5mm}
\end{figure}

  
 \begin{figure}[t]
\centering
\includegraphics[scale=0.25]{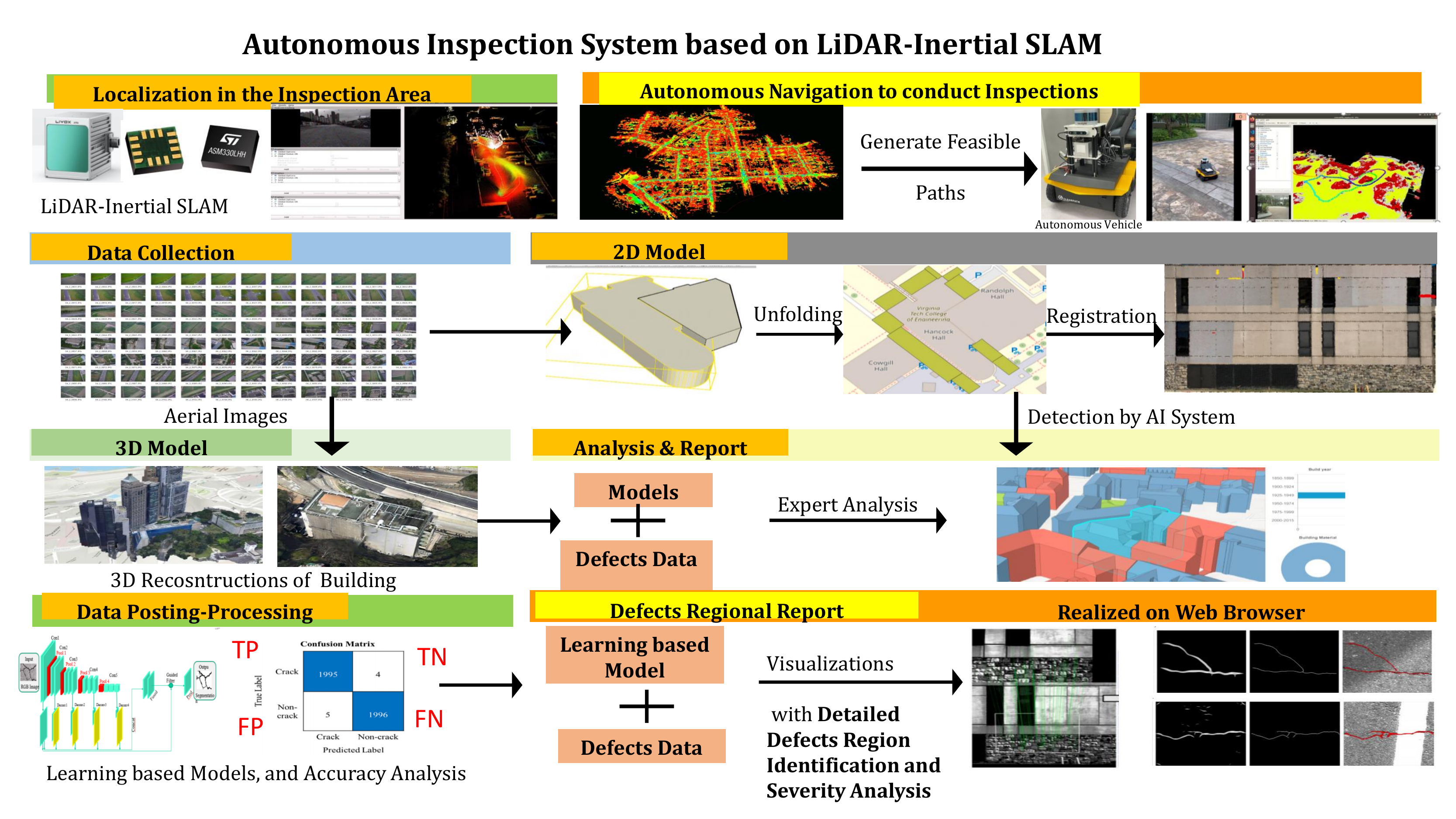}
\caption{Detailed illustration of the deployed system for UAV/UGV-based Inspections based on our proposed LVI-SLAM system.}
\label{fig_System_Inspection}
\vspace{-0.16cm}
\end{figure}


\subsubsection{Network Settings and Results}
 In our setting of the training of the network, we have adopted a learning rate of $5 \times 10^{-4}$ with a decay of 0.98 in each training epoch. And the training of our network lasts for 280 epochs. We implement the network framework using \textit{Pytorch}. Note that all our proposed loss function designs are merely required in the training stage. 
 \subsubsection{Integrating the Loop Closure into LVI-SLAM} Following LOAM, the edge features and planar features of the point clouds can be extracted. Let $n$ be the size of points. The roughness degree $R$ of each point can be calculated as:

\begin{equation}
    R= \frac{1}{n\cdot \| q_i\|}\| \sum_{j=1,j \neq i}^{n}(q_i-q_j)\|
\end{equation}
where $q_j$ is the adjacent points of the point $q_i$. The number of adjacent points is denoted as $n$. Then, a threshold $\beta$ of the roughness degree is set to categorize the points into either the edge point or the plane point. The points are classified as the edge points if the roughness degree is larger than $\beta$, and the plane points if the roughness degree is less than $\beta$. 

Then, we can do the state estimation by the optimization that minimize the distance of  edge feature points to their corresponding line and the distance of each planar feature points to their corresponding points. The final optimization objective function $f$ of the proposed SLAM system can be summarized as:
\begin{equation}
    f= \sum_{s}(y_1l_e^s)+ \sum_{t}(y_2l_p^t).
\label{eq_optimize}
\end{equation}

The $l_e^s$ is the distance of the \textit{s-th} edge feature point to its corresponding line. And the $l_p^t$ is the distance of the \textit{p-th} planar feature to its corresponding line. We adopt the balance weights $y_1$ and $y_2$ according to the LIO-SAM \cite{shan2020lio}.



\section{Experiments}

\begin{table}[t!]
\caption{The comparisons of localization accuracy and computational cost tested on the Hong Kong Science Park shown in Fig. \ref{fig_overall}, with the range of around 1000m $\times$ 150m (Left Value), and the tunnel environment as shown in Fig. \ref{fig_overall}, with the range of about 1200m $\times$ 90m (Right Value).}
\label{table_data}
\begin{center}
\scalebox{0.71}{\begin{tabular}{ccc}
\toprule

Methods & Computational Time (ms/frame) & Error (cm)\\
\toprule
Ours (\textit{LIO-Only})&46.58/46.96&8.75/8.86\\
Ours (\textit{VIO-Only})&47.62/46.28&10.91/9.18\\
Ours (\textit{LVI-SLAM w/o backend}))&61.86/62.03& 7.92/7.55\\
Ours (Full LVI-SLAM with Loop Closure)&83.36/83.27&7.05/6.92\\
Ours (Full LVI-SLAM)&77.23/76.52&5.87/5.69\\
\bottomrule
\end{tabular}}
\end{center}
\vspace{-3mm}
\end{table}

\begin{table}[t!]
\caption{Ablation studies of the \textit{LC-Net} on the Localization Performance with the LVI-SLAM experiments tested on the Hong Kong Science Park shown in Fig. \ref{fig_overall}, with the range of around 1000m $\times$ 150m (Left Value), and the tunnel environment as shown in Fig. \ref{fig_overall}, with the range of about 1200m $\times$ 90m (Right Value).}
\label{table_lc_ablation}
\begin{center}
\scalebox{0.96}{\begin{tabular}{cc}
\toprule
Methods in Comparisons & Average Localization Error (cm)\\
\hline
Ours (Full)& 5.87 / 5.69 \\
Ours (w/o $L_{ce}$)& 6.25 / 6.13\\
Ours (w/o $L_{reg}$)& 6.01 / 5.87\\
\bottomrule
\end{tabular}}
\vspace{-5mm}
\end{center}
\end{table}

In order to demonstrate the effectiveness of the proposed LVI-SLAM system, we have done extensive experiments in various of circumstances. We have utilized the Velodyne VLP-16 as our LiDAR sensor, and the orignial USB camera (with the price of 15 USD) as our visual sensor. All our proposed LVI-SLAM system is run on a Nvidia-Jetson-Xavier micro-computer (with the weight of less than 300 g). The control inner loop stabilizes the attitude of the UAV, while the outer loop tracks the position, velocity, and acceleration commands from the UAV motion planner. Among diverse control approaches, the composite nonlinear feedback (CNF) method for attitude stabilization and robust perfect tracking (RPT) method for trajectory tracking show great performance in inspections. We use a PX4-based flight micro-controller to implement the flight control law, and realize safe-corridor based autonomous motion planning strategy to automatically avoid obstacles when conducting inspection tasks. And we utilize the NVIDIA Jetson AGX Xavier (or the Jetson TX2 as an alternative) as the onboard mission computer. Communications of UAVs with other UAVs and the Ground Control Station (GCS) are done using WIFI. Also, we have constructed the autonomous navigated UGV system for inspections shown on the right side of Fig. \ref{fig_System_Inspection}. 


 \subsection{Quantitative Experimental Results}

 We have done extensive experiments to demonstrate the effectiveness of our proposed framework, the results are shown in  Table \ref{table_data} and Table \ref{table_lc_ablation}. From the first three rows in Table. \ref{table_data}, we can see that the tightly coupled front-end optimization of the LVI-SLAM has significant boosts on the localization accuracy. In addition, the overall performance drops without the factor-graph-based background optimization, which demonstrates that the back-end optimizations are also important to globally correct some localization errors. Moreover, without the proposed loop closure detection strategy, the localization performance also drops by a large margin, which demonstrates that the proposed loop closure network \textit{LC-Net} is also very significant for correcting some accumulated long-term drifts.  It can be demonstrated that our proposed modules are all significant for improving the localization accuracy, and they formed a robust and accurate LVI-SLAM system. It can be also demonstrated that our proposed LVI-SLAM system achieves the speed of more than 10Hz, which fulfills the requirements of real-time localization and mapping. From the Table \ref{table_lc_ablation}, we can see that both the classification head loss function and the regression head loss function contribute to the improvement of the global localization accuracy.

\subsection{Qualitative Experimental Results}

 In our setting of the training of the \textit{LC-Net}, we have adopted a learning rate of $5 \times 10^{-4}$ with a decay of 0.98 in each training epoch. And the training of our network lasts for 280 epoches. We implement the network framework using \textit{Pytorch}. We have done experiments in the outdoor road case such as Mulran benchmark with mechanical LiDAR and the harbor case locally at Hong Kong with solid-state LiDAR. Also, we have tested under various of complicated circumstances at Hong Kong science park and the tunnel environments as shown in Fig. \ref{fig_overall}. From the first row, it can be demonstrated that our propose network can precisely detect the same scan. Also, our proposed network can also provide very accurate and precise predictions of the relative angle between two different scans. 
 \begin{table}[t]
\caption{The comparisons results of the loop closure test success rate at the tunnel environments}
\label{table_loop_close}
\begin{center}
\scalebox{1.00}{\begin{tabular}{ccc}
\toprule

Methods & Success & Failure in Loop Closure\\
\hline
Ours (w/o \textit{LC-Net})&4&3\\
Ours (Full w/ \textit{LC-Net})&7&0\\
LPD-net &5&2\\
LEGO-LOAM &2&5\\ 
LIO-SAM &3&4\\
\bottomrule
\end{tabular}}
\end{center}
\vspace{-5mm}
\end{table}

\subsection{Real-Site Robot Navigation Integrated LVI-SLAM System}
 It has been demonstrated by experiments that our proposed LVI-SLAM system has great capacity to generate high-quality map, which is of great help to conduct autonomous motion planning to generate feasible path.  The left shows the 3D mapping results in the tunnel environment, and the right shows the defects analysis and the tunnel environmental analysis based on the LiDAR 3D mapping results. Our proposed LVI-SLAM system can be integrated seamlessly with the motion planning approaches to realize fully autonomous navigation for both the unmanned aerial vehicle and unmanned ground vehicle in the narrow corridor and complex tunnel environments.    
\section{Conclusions}
In this work, we have proposed an integrated improved LiDAR-Visual-Inertial simutaneous localization and mapping system. It can be demonstrated that our proposed LiDAR-Visual-Inertial SLAM system show great performance and robustness in both indoor and outdoor circumstances with mechanical LiDAR or solid-state LiDAR. We have proposed the projection-based method for find the 2D-3D alignment and formulated it into the iterative Kalman filter-based fusion of visual and LiDAR information. Also, we have proposed the factor graph-based optimizations for the back-end optimizations of our proposed LVI-SLAM system. Finally, we have proposed the loop closure detection network to further enhance the global localization and mapping performance. Extensive experiments at the science park, road scenarios the tunnel environments real-site at Hong Kong demonstrate that the proposed LVI-SLAM system has superior performance in both efficiency and accuracy.

\addtolength{\textheight}{0cm}   





\bibliographystyle{IEEEtran}
\bibliography{references}

\end{document}